\pdfoutput=1

\documentclass[11pt]{article}

\usepackage[final]{acl}%

\usepackage{times}
\usepackage{latexsym}

\usepackage[T1]{fontenc}

\usepackage[utf8]{inputenc}

\usepackage{microtype}

\usepackage{inconsolata}

\usepackage{graphicx}
\usepackage{authblk}
\usepackage{longtable}

\newcommand*\samethanks[1][\value{footnote}]{\footnotemark[#1]}

\title{Balancing Quality and Variation: Spam Filtering Distorts Data Label Distributions}

\author[1]{Eve Fleisig\thanks{Equal contribution; order determined by coin flip.}}
\author[2]{Matthias Orlikowski\samethanks}
\author[2]{Philipp Cimiano}
\author[1]{Dan Klein}
\affil[1]{UC Berkeley, \texttt{\{efleisig,klein\}@berkeley.edu}}
\affil[2]{Bielefeld University, \texttt{\{morlikowski,cimiano\}@techfak.uni-bielefeld.de}}

\begin{document}
\maketitle
\begin{abstract}
For datasets to accurately represent diverse opinions in a population, they must preserve variation in data labels while filtering out spam or low-quality responses. How can we balance annotator reliability and representation? We empirically evaluate how a range of heuristics for annotator filtering affect the preservation of variation on subjective tasks. We find that these methods, designed for contexts in which variation from a single ground-truth label is considered noise, often remove annotators who disagree \textit{instead} of spam annotators, introducing suboptimal tradeoffs between accuracy and label diversity. We find that conservative settings for annotator removal (<5\%) are best, after which all tested methods increase the mean absolute error from the true average label. We analyze performance on synthetic spam to observe that these methods often assume spam annotators are more random than real spammers tend to be: most spammers are distributionally indistinguishable from real annotators, and the minority that are distinguishable tend to give relatively fixed answers, not random ones. Thus, tasks requiring the preservation of variation reverse the intuition of existing spam filtering methods: spammers tend to be \textit{less} random than non-spammers, so metrics that assume variation is spam fare worse. These results highlight the need for spam removal methods that account for label diversity.

\end{abstract}

\section{Introduction}

Because spam responses are common on crowdsourcing sites, researchers need reliable ways to filter out low-quality data. Many of these methods aim to find annotators with unusual labeling behavior. However, a growing body of work has found that information from annotators with minority opinions can be a valuable source of information, since this disagreement helps to understand variability in the opinions of a population, identify cases where some annotators may be better- or worse-informed, or reveal ambiguity in the task. How can we preserve the opinions of annotators who disagree, while still removing spam annotations?

We examine the effects of applying several common methods for discounting spam annotators based on their labeling behavior. We test three of these methods---MACE \cite{hovy-etal-2013-learning}, CrowdTruth \cite{aroyo_three_2014}, and inter-annotator agreement metrics---on relatively subjective tasks and analyze effects on variability in the filtered data. We find that, although many methods are near-indistinguishable in terms of their \textit{accuracy} at classifying spam annotators, some are far more likely to remove non-spam annotators who disagree. Furthermore, we find that under most tested methods, removing more annotators degrades the variety of opinions expressed, without improving accuracy at removing spam annotators; thus, these methods seem most effective only when a very low number of annotators are removed.

We also find that assumptions about the distribution of spam annotations can hinder the effectiveness of these methods. We examine performance on synthetic distributions of spam annotations to analyze whether these methods effectively remove spam annotations, or simply remove annotations farther from the mean. Performance on synthetic spam indicates that most methods perform far better for random spam (e.g., randomly clicking answers) than fixed spam (e.g., always answering “No”). Yet true spammer behavior exhibits the opposite trend: most spammers are distributionally similar to high-quality annotators; the minority that can be reliably identified tends to have fixed spamming behavior. As a consequence, methods that perform poorly on fixed spam tend to also perform poorly on real spam.

Despite the existence of spam removal methods that use attention checks or metadata (e.g., time spent on task), filtering based on labeling behavior remains common practice \cite{klie-etal-2024-analyzing}; thus, weaknesses in these methods risk affecting a wide range of common machine learning tasks. Our results indicate that spam detection for subjective problems flips the assumptions of existing methods: spam annotators are often less random than non-spam ones.
Thus, attempts to remove spam can backfire by instead removing annotators with minority opinions who are not spammers. As a result, existing methods work best when only low percentages of annotators are removed based on their labeling behavior. When over-filtering for spam, these methods risk distorting the distributions of labels.

\section{Related Work}

Our study connects work on \emph{spammer detection and label aggregation}, which focuses on the quality of ground truth,  to work on \emph{subjectivity and variation in annotation}, which questions the idea of a single truth but rarely engages with quality concerns.

\paragraph{Defining Spammers in Annotation.}
Drawing a conceptual boundary between spammers and genuine annotators is complex. Definitions vary on what range of intentional, inattentive, or low-effort behaviors should be filtered out, and on whether spammers are posited as too random or too fixed. \citet{buchholz_crowdsourcing_2011} highlight that spammers are incentivized to earn more money faster, leading them to ignore task instructions or participation requirements. \citet{rothwell_controlling_2015} argue that spammers act with intention, unlike other types of low-quality annotators, and show \emph{repeated patterns} in an attempt to complete tasks fast. In contrast, \citet{raykar_eliminating_2012} posit that spammers \emph{assign labels randomly}, because they do not follow labeling criteria, skip reading the instances or might use automation. \citet{gadiraju_understanding_2015} present a nuanced taxonomy of annotator types and underscore that genuine annotators' behavior might overlap with spammers, e.g., failing attention checks for innocuous reasons.
The datasets used in our study excluded annotators if they failed data quality checks combining multiple sources of information, thus following a wider definition of spam \cite[][see Section \ref{datasets}]{aroyo2023dices,huang-etal-2023-incorporating}.
Any definition of ``spammer'' includes or excludes different subsets of annotators; these ambiguous boundaries suggest that different subsets of spammers may exhibit different behaviors, potentially raising challenges in distinguishing spammers from non-spammers.

\paragraph{Spammer Detection and Gold Label Aggregation.}
\label{sec:related_work_quality}
Data quality and questionable trust in non-expert raters are longstanding problems in crowdsourced annotation  \cite{snow-etal-2008-cheap}. Attempts to improve data quality may modify tasks to attract less spam before data collection \cite{eickhoff_increasing_2013} or use quality control afterwards \cite{difallah_etal_2012_mechanical_cheat}. Methods for a posteriori detection of low-quality raters and spammers often use intrinsic metrics based on the labeling behavior itself \cite{buchholz_crowdsourcing_2011}. Intrinsic metrics used for spammer detection include clustering on a post-processed annotation matrix \cite{traganitis_identifying_2021}, rater similarity and agreement scores \cite{ak_spammer_2021}, or analyzing sequential spamming behaviors \cite{ba_data_2024}, among others \cite{ipeirotis_quality_2010,raykar_eliminating_2012,gadiraju_understanding_2015}. Other methods analyze labeling behavior with the goal of \emph{aggregating to the true label} while accounting for varying annotator reliability. \citet{dawid_maximum_1979} model annotator error rates to estimate the true labels and are foundational to many subsequent aggregation methods \cite{whitehill_whose_2009, welinder_multidimensional_2010}, including in NLP \cite{wiebe-etal-1999-development}. \citet{passonneau-carpenter-2014-benefits} present a probabilistic variant of the Dawid \& Skene model, and many other extensions of this basic model exist \cite[][]{paun-etal-2018-comparing,paun_statistical_2022}. In particular, \citet{hovy-etal-2013-learning} present MACE, a probabilistic model tailored towards estimating annotator competence by modeling spamming behaviors. In contrast, CrowdTruth, a non-probabilistic paradigm,  derives quality metrics from vector space representations of annotators, annotated examples and annotations \cite{aroyo_three_2014,CrowdTruth2}. We evaluate MACE and CrowdTruth as they underwent widespread adoption in NLP and have reference implementations available (see Sections \ref{sec:methods_mace} and \ref{sec:methods_crowdtruth}).

\paragraph{Subjectivity and Variation in Annotation.}
There is a growing body of work researching informative disagreement, diversity of perspectives, and label variation in human annotation~\citep{plank-2022-problem, leonardelli-etal-2023-semeval,sandri-etal-2023-dont,frenda_perspectivist_2024,fleisig-etal-2024-perspectivist}. These studies agree that aggregating labels into a single truth is an oversimplification for many tasks~\citep{aroyo_truth_2015,uma-2021-survey, basile-etal-2021-need} and might not represent perspectives fairly~\citep{nlperspectives-2022-perspectivist}. Instead, studies release annotator-level labels \cite{prabhakaran-etal-2021-releasing} to enable alternative approaches, such as modeling individual annotators' rating behaviors \cite{fleisig-etal-2023-majority,orlikowski-etal-2023-ecological,heinisch-etal-2023-architectural,orlikowski-etal-2025-beyond}. Our work is motivated by studies on \emph{rating distributions in a given population} as an alternative to single ground truth prediction \cite{sorensen_etal_2025_pluralistic,meister-etal-2025-benchmarking}. Among these, \citet{prabhakaran-etal-2024-grasp} study systematic disagreement using similar metrics to ours, but on the level of demographic subgroups. 
In this context, the issue of how capturing labeling variation intersects with annotation quality is largely unexplored. One exception is VariErr \cite{weber-genzel-etal-2024-varierr}, an annotation methodology to differentiate between annotation errors and plausible variation in annotation. In contrast, we study properties of methods that determine annotator reliability, not individual annotation errors.

\section{Datasets}
\label{datasets}
We selected two datasets for the basis of our experiments: DICES 350 \cite{aroyo2023dices} and \citet{huang-etal-2023-incorporating}'s survey of Amazon Mechanical Turk workers. We present each dataset's statistics and discuss our dataset selection process below.

\paragraph{DICES-350}

DICES-350 \cite{aroyo2023dices}, a harmful language dataset, consists of 43,050 annotations on a 3-point scale across 350 items, each of which was labeled by every participant. 123 annotators participated, of whom 19 annotators were labeled as spam (15\% of annotators). 

\paragraph{MTurk}

From \citet{huang-etal-2023-incorporating}'s survey, we used 16 questions on a 7-point scale, each of which was answered by every participant, for a total of 3,312 annotations. 207 annotators participated, of whom 40 were labeled as spam (19\% of annotators).

\paragraph{Dataset selection.}

Our experiments require datasets that retain (a) responses from multiple annotators per question, permitting measurement of disagreement statistics, and (b) responses from known spammers. Despite increasing availability of annotator-level data,\footnote{For example, \url{https://github.com/mainlp/awesome-human-label-variation}} most public datasets do not include spammer information. For papers that report spammer removal, we contacted authors for access to the unfiltered datasets, but spammer responses are regularly lost over time \cite[e.g.,][]{buchholz_crowdsourcing_2011,dumitrache_false_2017,paun-etal-2018-comparing}. Even for published data, maintaining data access is not always possible; some datasets with verified spammer information were no longer available \cite[e.g.,][]{soberon_measuring_2013, gadiraju_understanding_2015}. Similarly, many studies on spammer detection evaluate only on downstream performance or exclusively use synthetic data, so they do not provide metadata on known natural spammers \cite[e.g.,][]{raykar_eliminating_2012,ak_spammer_2021}. See Appendix \ref{sec:appendix-dataset-selection} for details on all 22 considered datasets, including spammer metadata and data availability. In summary, DICES-350 and the MTurk survey are, to our knowledge, the only available datasets meeting our criteria. Nevertheless, these datasets do represent two representative use cases in which preserving rater variation is essential. DICES-350 collects annotations on AI safety to preserve variation on diverse perspectives regarding high-stakes topics; the MTurk dataset polls workers on their personal opinions about their crowdwork experiences in order to  understand the range of opinions of the community.

\section{Methods for Spammer Detection}

We study a number of established methods and baselines to calculate scores of annotator reliability. To perform spammer detection, we rank annotators using each method's reliability score and identify the $k$ lowest-scoring annotators as spammers for a given value of $k$.

\subsection{Multi-Annotator Competence Estimation (MACE)}
\label{sec:methods_mace}
MACE \cite{hovy-etal-2013-learning} is based on a probabilistic model of annotation. We highlight a few aspects of MACE that are important to our study and refer to the original paper for full details. The model includes a parameter $\theta$ for each annotator which encodes the probability that they give the true answer (\emph{competence}). Specifically, for each instance $i$ and annotator $j$, the binary variable $S_{ij}$ indicates whether an annotator is spamming. $S_{ij}$ is drawn from a Bernoulli distribution with parameter $1 - \theta_j$. If the annotator is spamming, i.e., $S_{ij} = 1$, then the assigned label $A_{ij}$ is sampled from a multinomial distribution with a parameter vector $\zeta_j$ that encodes each annotator's spamming strategy. Otherwise, if $S_{ij} = 0$, the model assumes that the annotator simply assigns the correct label---an intentional simplification to focus on modeling spam behavior. Only the annotations $A_{ij}$ are observed; the other parameters are inferred when updating the model from data.

Usually, when applying MACE for label aggregation, the model would weigh all annotations to estimate the correct labels without discarding specific annotators. But the learned parameters can also be used to identify spamming annotators: the competence $\theta$ correlates more strongly than agreement measures with an annotator’s fraction of correctly annotated examples \cite{hovy-etal-2013-learning} and both learned annotator parameters ($\theta$, $\zeta$) were shown to encode characteristic spamming behaviors \cite{paun-etal-2018-comparing}. Consequently, other studies have used MACE to exclude spammers during dataset construction based on an empirically chosen threshold for competence \cite{pei-jurgens-2023-annotator}. In our experiments, we also use the competence parameter to score annotators.

\subsection{CrowdTruth}
\label{sec:methods_crowdtruth}

The CrowdTruth framework \cite{aroyo_three_2014} computes several interdependent quality metrics that use vector representations of annotations to measure disagreement and ambiguity, including a worker quality score. The metrics follow the aim of ambiguity-aware label aggregation, so that, for example, disagreement on ambiguous instances discounts worker quality less. The worker quality score (WQS) for an annotator $i$ is computed as the product of two other scores $WQS(i) = WUA(i) \cdot WWA(i)$, the worker-unit agreement (WUA) and the worker-worker agreement (WWA). Conceptually, WWA measures how similar a given worker's annotations are to other workers, weighted by the workers' quality and the instances' ambiguity. WUA measures how much a worker agrees with the aggregate label over all their annotated instances, weighted by the instances' ambiguity. (See Appendix \ref{sec:appendix_crowdtruth} for details on how these metrics are computed.)

The CrowdTruth metrics were explored on various tasks \cite{dumitrache_false_2017,dumitrache_crowdsourcing_2018} and have been explicitly used for spammer removal \cite{dumitrache_empirical_2021}. In a related study, \citet{soberon_measuring_2013} report an accuracy of $0.88$ for removing spam annotators using CrowdTruth metrics. In our experiments, we use the worker quality score (WQS) to score annotators.

\subsection{Cohen's Kappa}

As a representative example of using inter-annotator agreement metrics to filter annotators, we compute each annotator's pair-wise agreement as measured by Cohen's kappa \cite{cohen_weighted_1968} with each other annotator. We then use the averaged agreement to score annotators.

\subsection{Random Baseline}
We assign scores to annotators (from 0.0 to 1.0) by drawing from a uniform distribution. 

\section{Results}
We applied MACE, Crowdtruth, the Cohen's kappa filter, and a random baseline on both datasets, as the threshold for number of annotators removed increases. For studies in spammer detection and gold label aggregation (see Section \ref{sec:related_work_quality}) the primary metric to optimize is downstream classification performance, often based on synthetic spam annotations, whereas we focus on tasks where preserving labeling variation is key. We measured the change in standard deviation, entropy, and accuracy of spammer detection for the DICES-350 and Mturk datasets. Standard deviation is computed over all labels. Entropy is computed for the label distribution of each instance and then averaged. We also measured the KL-divergence and mean absolute error of the filtered labels from the labels of non-spammers.

\begin{figure}[h]
  \centering
  \includegraphics[width=0.9\linewidth]{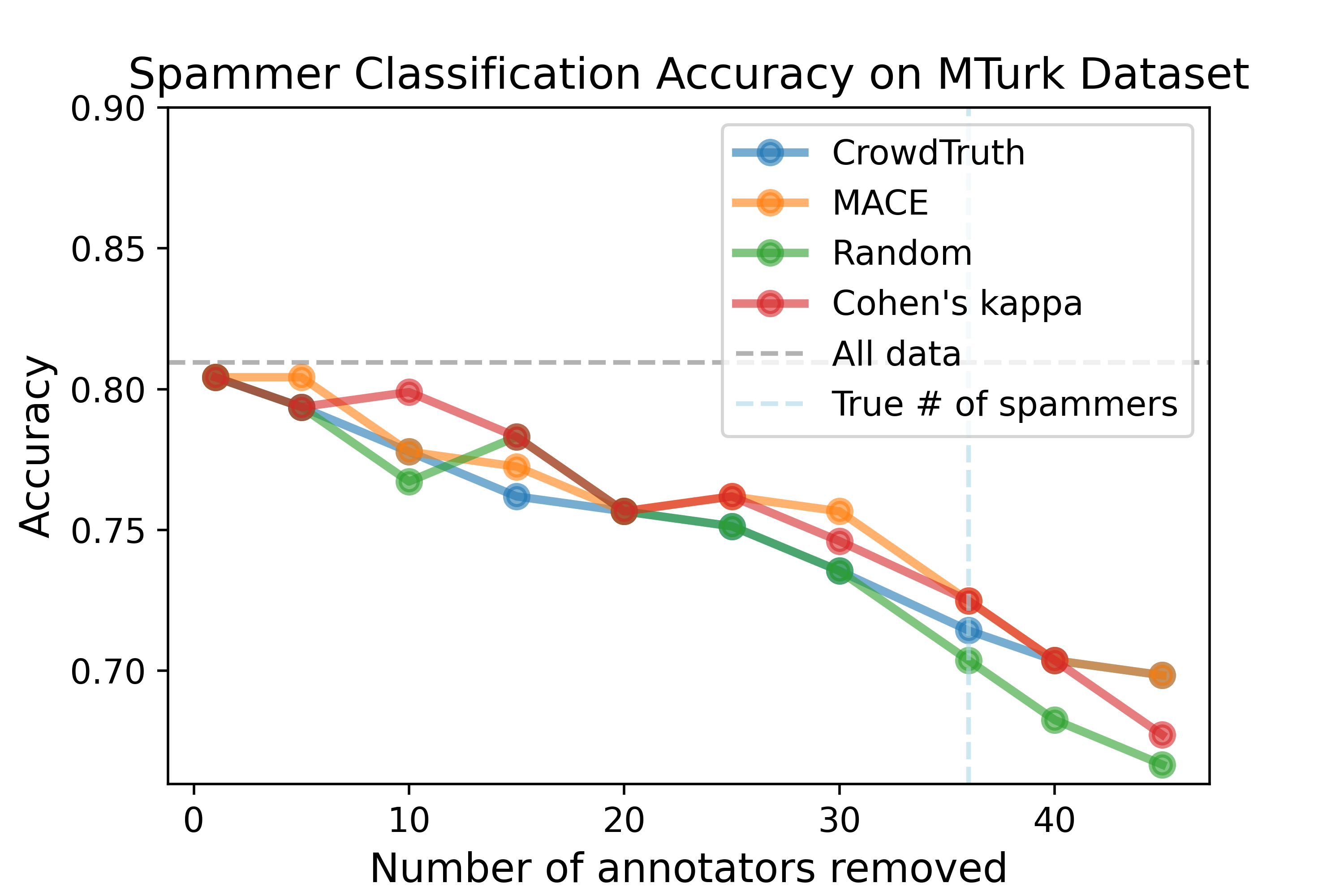}
\includegraphics[width=0.9\linewidth]{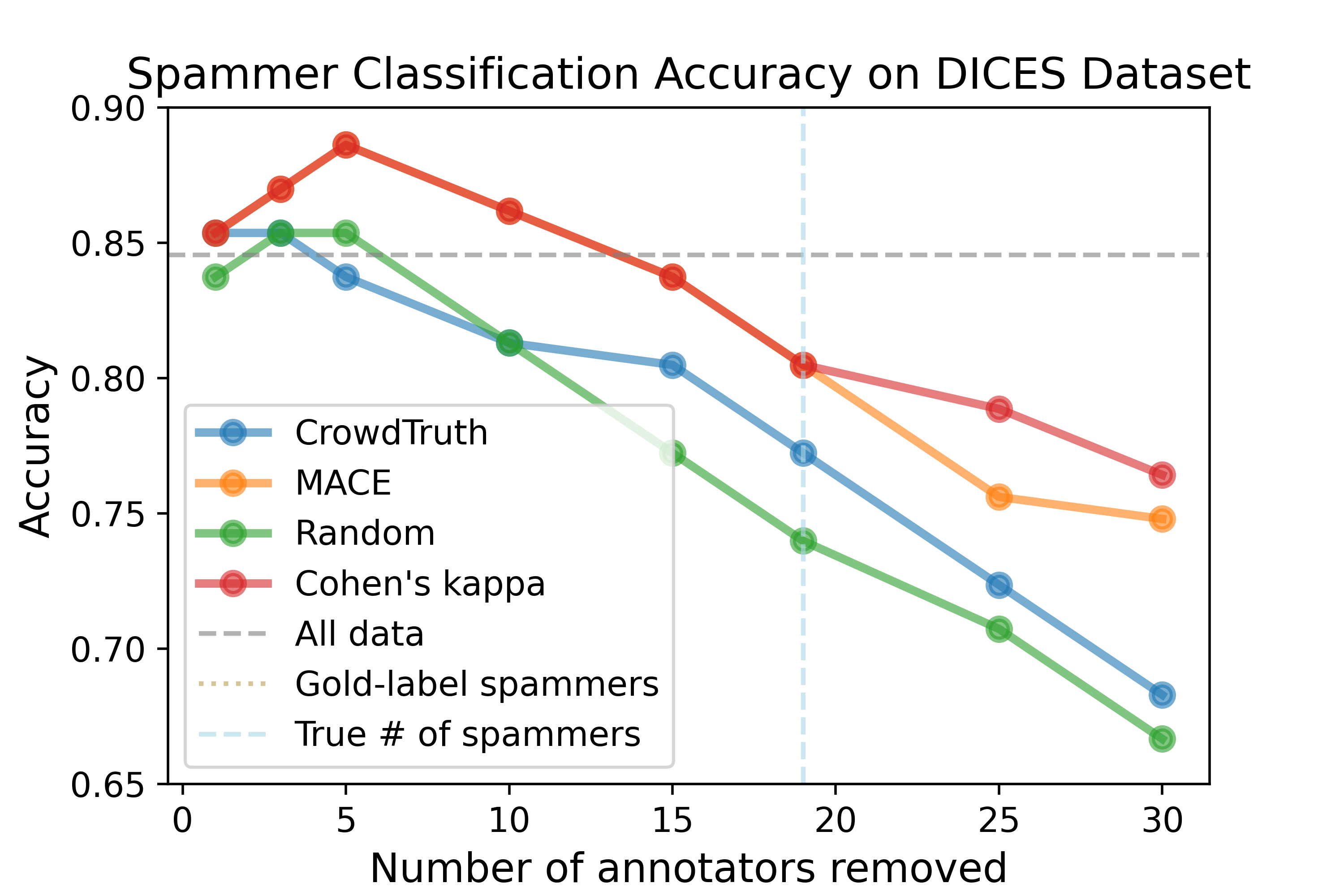}
  \caption{Across methods, increasing the number of removed annotators gradually decreases the accuracy of spam classification when over 2-4\% of annotators are removed. Cohen's kappa and MACE increase the spam classification accuracy up to 4\% of annotators removed on DICES;  otherwise, the spam classification accuracy rarely rises above the baseline of not removing any annotators. The blue line indicates the true number of  spammers in the data; the gray line indicates the baseline classification accuracy before removing any spammers.}
  \label{fig:accuracy}
\end{figure}

\subsection{Accuracy vs. Preserved Variation for Spam Detection}

Across methods, increasing the number of removed annotators gradually decreases the accuracy of classifying annotators as spammers (Figure \ref{fig:accuracy}, top). For the MTurk dataset, the accuracy of spam classification never rises above the accuracy of not removing any annotators. For the DICES dataset,  Cohen's kappa and MACE outperform removing zero annotators when <10\% of annotators are removed, while CrowdTruth and random removal quickly fall below baseline accuracy (Figure \ref{fig:accuracy}, bottom). The best accuracy is achieved when only focusing on the lowest-scoring annotators (lowest 2-4\%). 

\begin{figure}
  \centering
\includegraphics[width=0.9\linewidth]{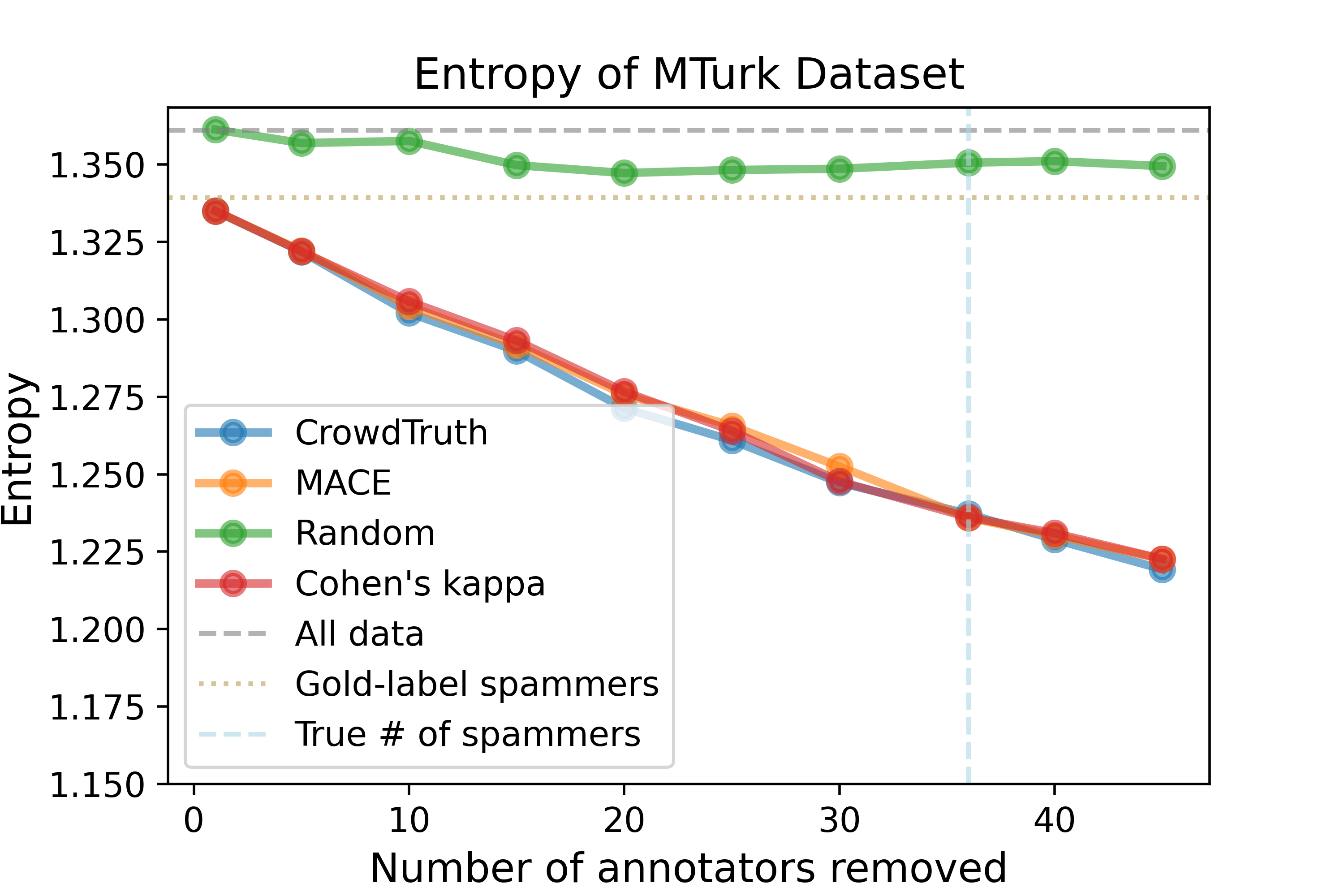}
\includegraphics[width=0.9\linewidth]{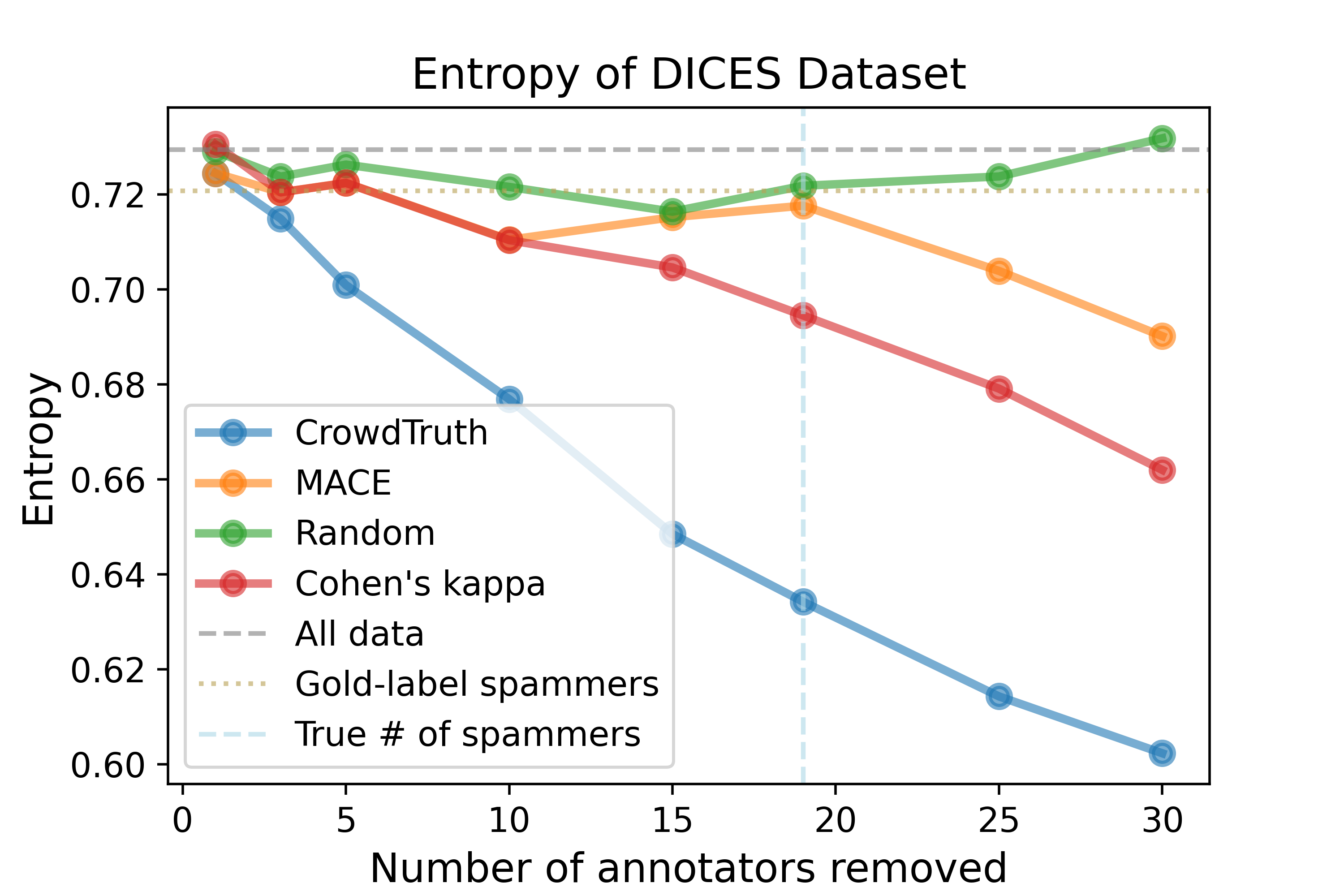}
  \caption{Entropy of each instance's label distribution, averaged over all instances. Most methods decrease the entropy of the dataset as more raters are removed. CrowdTruth especially decreases the entropy.}
  \label{fig:entropy}
\end{figure}

\begin{figure}
  \centering
  \includegraphics[width=0.8\linewidth]{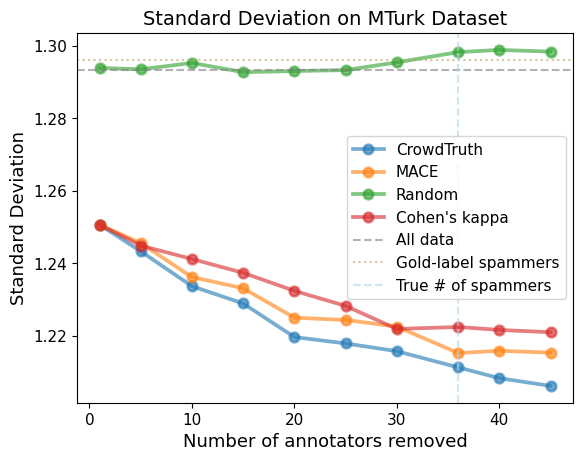}
\includegraphics[width=0.8\linewidth]{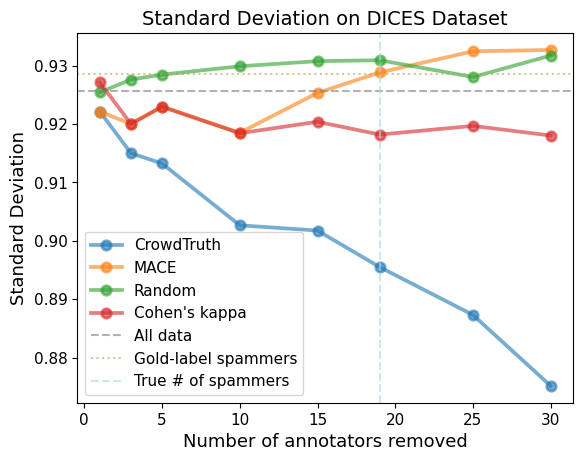}
  \caption{On the MTurk dataset, all methods except random removal decrease the standard deviation of the dataset. Among the tested methods, CrowdTruth decreases the standard deviation most.}
  \label{fig:stddev}
\end{figure}

We also measured the change in entropy and standard deviation of the filtered dataset, finding that these methods typically reduce variance in the distribution of annotator opinions, discarding information about annotator disagreement (Figures \ref{fig:entropy} and \ref{fig:stddev}). Except for the random baseline, the tested methods generally decrease the entropy of the distributions as more raters are removed. This is especially true of CrowdTruth, which quickly decreases the entropy; MACE and Cohen's kappa also decrease the entropy to a lesser extent. CrowdTruth also consistently decreases the standard deviation of the data. MACE and Cohen's kappa decrease the standard deviation on the MTurk dataset, but not on DICES.

\begin{figure*}
  \centering
  \includegraphics[width=0.45\linewidth]{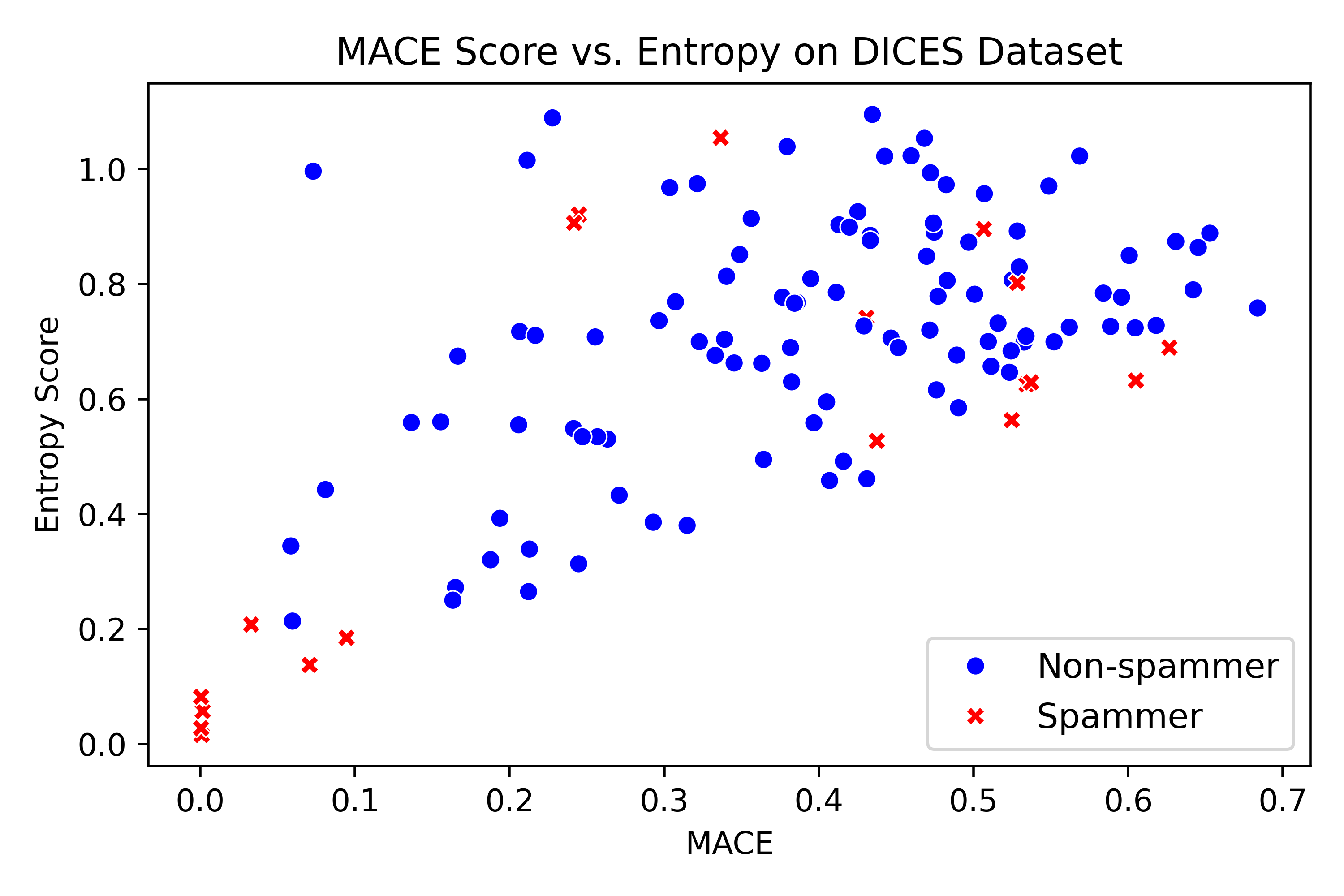}
\includegraphics[width=0.45\linewidth]{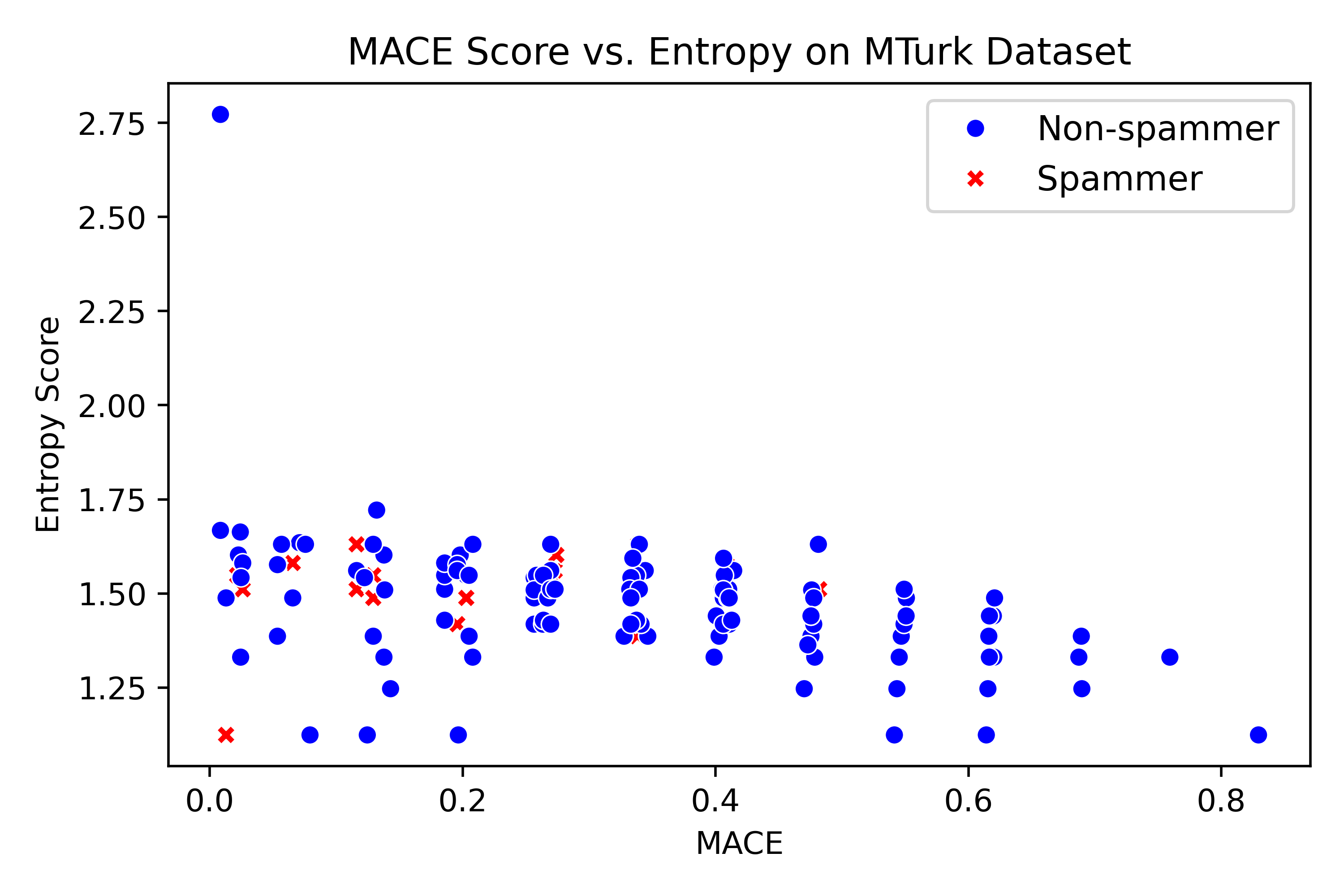}
\includegraphics[width=0.45\linewidth]{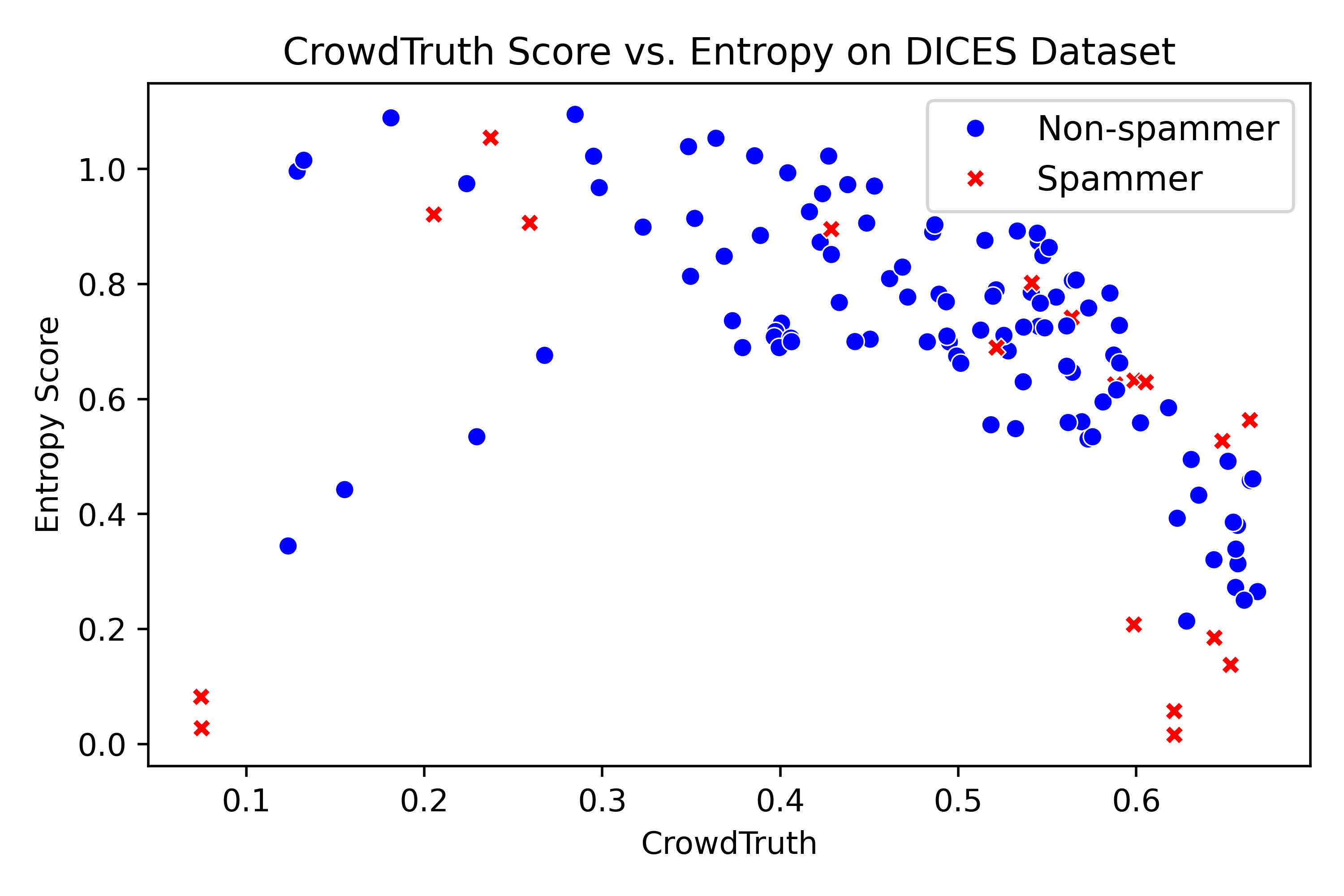}
\includegraphics[width=0.45\linewidth]{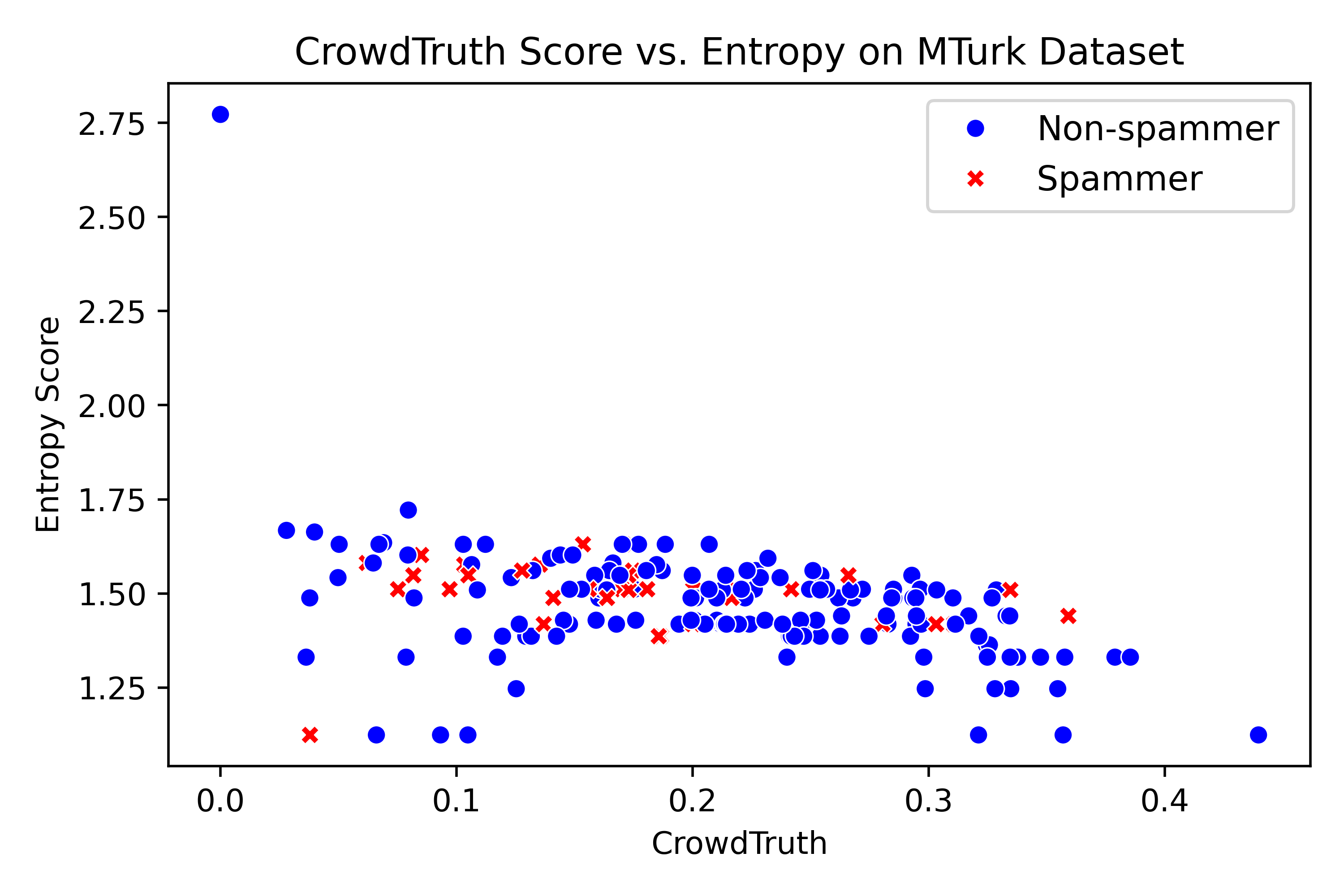}
  \caption{Entropy of each annotator's labeling distribution over all instances vs. score under filtering metrics (CrowdTruth and MACE). While many spam annotators are indistinguishable from non-spam ones under these metrics, those that are often have very low entropy: they are \textit{less} random than non-spam annotators, not more.}
  \label{fig:scatter-entropy}
\end{figure*}

\begin{figure}
  \centering
  \includegraphics[width=0.9\linewidth]{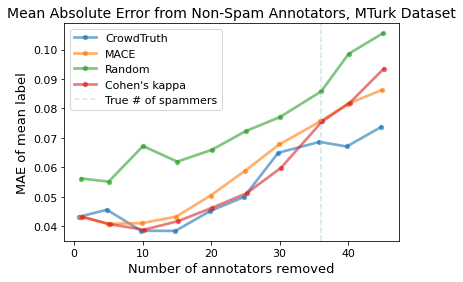}
\includegraphics[width=0.9\linewidth]{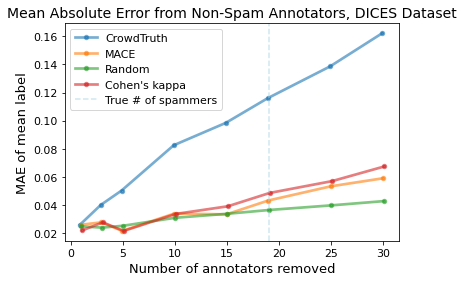}
  \caption{Mean absolute error of filtered ratings. Difference between average label on an example of non-spam annotators and filtered annotators, then averaged across examples.}
  \label{fig:mae}
\end{figure}

\begin{figure}
  \centering
  \includegraphics[width=0.85\linewidth]{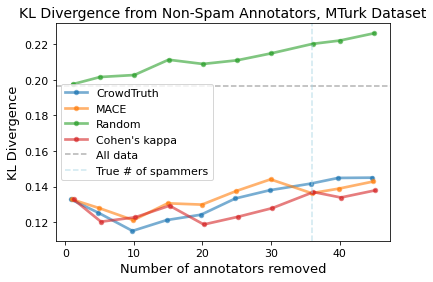}
\includegraphics[width=0.85\linewidth]{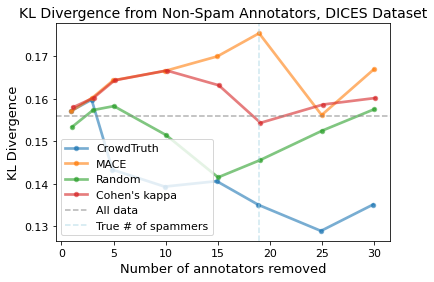}
  \caption{KL-divergence of filtered annotators per data item vs. true non-spam annotators, averaged across examples.
  }
  \label{fig:kldiv}
\end{figure}

To understand whether these methods affect how well the filtered datasets represent the true distribution of non-spam annotators' ratings, we also measured the mean absolute error (MAE) per example between the filtered annotators and the true non-spam annotators (i.e., the difference between their average labels on a given example; Figure \ref{fig:mae}) and the KL-divergence between the filtered and non-spam annotators (Figure \ref{fig:kldiv}). All tested methods eventually increase the mean absolute error, indicating that the mean label of the filtered data drifts away from that of the true non-spam annotators as more labels are removed. However, the extent of this varies  by method and dataset: on the MTurk dataset, all non-random methods have relatively little change in MAE when <5\% of annotators are removed, but increases after that; on the DICES dataset, CrowdTruth worsens the MAE much faster than other tested methods. The KL divergence remains relatively steady, but eventually increases on the MTurk dataset for all non-random methods, and fluctuates widely across methods on the DICES dataset.

Why might these methods fail to capture all spammers? Comparing the entropy of the responses given by each annotator with their scores under these metrics helps to understand where the assumed spammer behavior, as modeled by these metrics, differs from the spammer behavior seen in practice (Figure \ref{fig:scatter-entropy}). Most annotators lie well within the distribution of non-spammers in terms of entropy, MACE score, and CrowdTruth score. However, a subset of annotators are distinguishable as spammers (best seen on the DICES dataset) because they have especially low entropy. MACE captures many of these annotators, but CrowdTruth only captures some of them, perhaps explaining the difference in these metrics.

Since a cluster of spam annotators that can be reliably distinguished tends to have especially fixed behavior, perhaps models perform best at capturing spam if they can identify annotators with unusually fixed annotation patterns. To investigate this, we next studied model performance using synthetic spam.

\subsection{Synthetic Spam Analysis}
\label{sec:results-synthetic-spam}
To understand what factors affect spam detection methods' accuracy at classifying spam, and propensity to misclassify annotators who disagree as spam, we experiment with several kinds of synthetic data. \textit{Random spam} experiments simulate spam annotators whose annotations are random; \textit{fixed spam} experiments simulate spam annotators who always give the same answer, which is set to the mode response for the dataset.

\begin{figure*}
  \centering
  \includegraphics[width=0.4\linewidth]{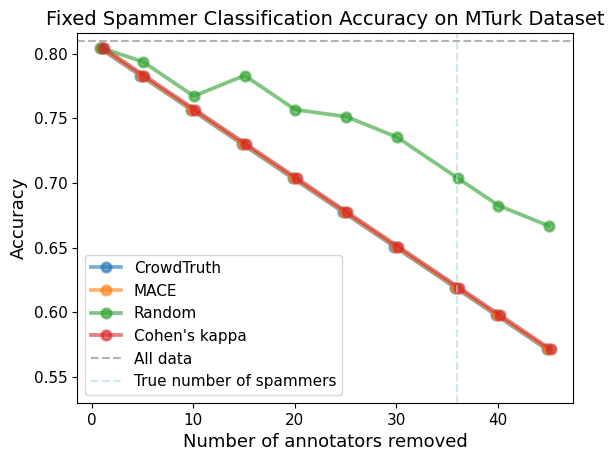} \includegraphics[width=0.41\linewidth]{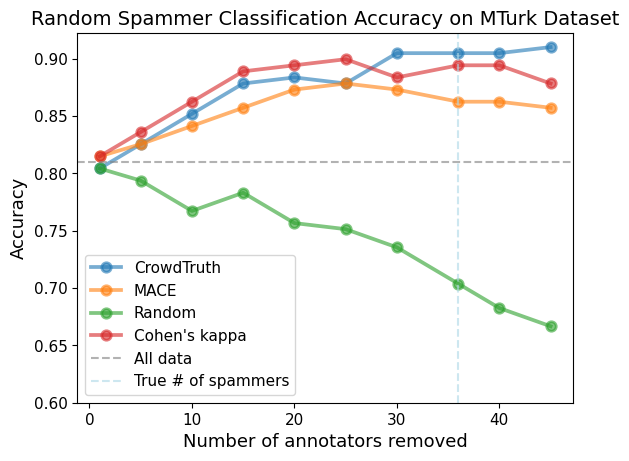}
\\
\includegraphics[width=0.39\linewidth]{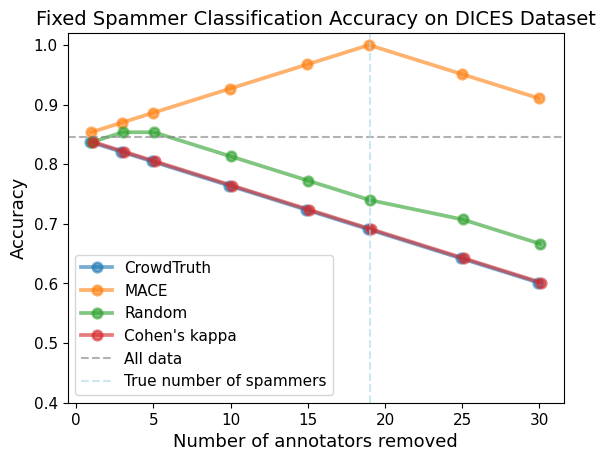}\includegraphics[width=0.41\linewidth]{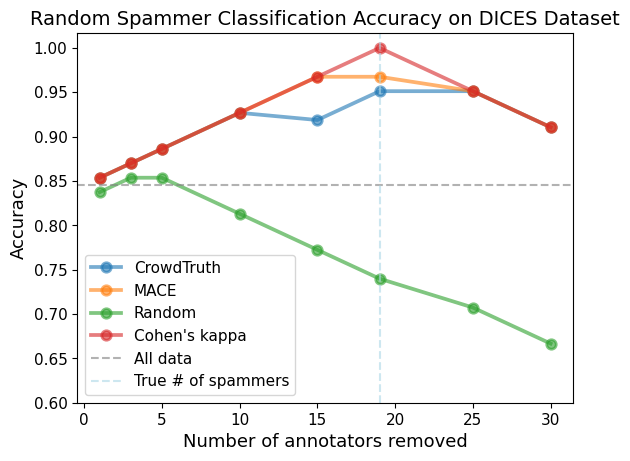}
  \caption{Accuracy with fixed-spam and random-spam synthetic annotators. For DICES, MACE performs best on fixed data; the other methods universally struggle. For random spam, all methods outperform the baseline, with Cohen's kappa performing optimally on DICES.}
  \label{fig:fixed-accuracy}
\end{figure*}

\paragraph{Fixed spam.} Because these methods tend to filter out annotators who are farther from the mean, most of them struggle to filter out annotators whose behavior is fixed to the mode value (e.g., answering "No" to every question). MACE performs much better than the other methods on fixed spam for DICES, but all methods are worse than the baseline for the fixed spammers on the MTurk data (Figure \ref{fig:fixed-accuracy}, left). 
MACE's higher accuracy on DICES  can partially be explained by how well the method can capture fixed spamming behavior given how it is set up (see Section \ref{sec:methods_mace}): A spammer would have low competence $\theta$, so that the assigned label is frequently sampled from the annotator's spamming strategy $\zeta$. As the spammer assigns always the same label, the parameter vector would encode high probability for that particular label and low probability for all others. In contrast, CrowdTruth factors in ambiguity but is ultimately based on agreement (see Section \ref{sec:methods_crowdtruth}). As a spammer who always assigns the mode label can score relatively high agreement in subjective tasks with stronger labeling variability, fixed spam annotators are not filtered out by CrowdTruth. This result about agreement for fixed spam is in line with the accuracy scores by Cohen’s kappa filtering, which are identical to CrowdTruth. Notably, this observation does not transfer to real spammer behavior (Figure \ref{fig:accuracy}), where Cohen’s kappa and MACE (but not CrowdTruth) similarly perform above baseline accuracy.

MACE's poor accuracy on fixed spam for MTurk is surprising given its perfect accuracy on DICES. This result is likely caused by answers in the MTurk dataset mostly following a normal distribution with the same mode, so that MACE overestimates the competence of fixed spammers (in contrast to DICES data; see Appendix \ref{sec:appendix-mace-fixed-spam}). 

Because the spammers all give the same ratings, we expect accurate spam classification to increase the standard deviation and the entropy, as happens for MACE on DICES; by contrast, Crowdtruth and Cohen's kappa filtering on DICES (and MACE on MTurk) decrease the standard deviation and the entropy without ever increasing spam classification accuracy above the baseline (Appendix \ref{sec:appendix}).

\paragraph{Random spam.} On the random data (Figure \ref{fig:fixed-accuracy}, right), CrowdTruth, MACE, and Cohen's kappa have similar accuracies noticeably above the baseline (peaking when the number of annotators removed equals the number of spam annotators). This suggests that random spam is closest to the spam behavior for which these methods work optimally.

In this case, we expect accurate spam classification to decrease the entropy, which indeed happens for both datasets across methods (Appendix \ref{sec:appendix}); the standard deviation also decreases for MTurk, and is more random for DICES, likely because DICES has a smaller set of possible answer values.

Together, these results suggest that real spam annotators are less random than the imagined spammer behavior under CrowdTruth and inter-annotator agreement filtering. This makes these methods vulnerable to removing annotators who are further from the mean rather than actual spammers. MACE, which is more robust to filtering out fixed spammers, also performs better at filtering out real spammers.

\section{Discussion and Conclusion}
 
\paragraph{Spam detection for subjective problems flips model assumptions: spam annotators are often \textit{less} random than non-spam ones.} Longstanding paradigms of annotation, focused on improving downstream model accuracy under the assumption of a single ground truth, often assume that disagreement indicates low-quality annotations. However, in problems where disagreement is expected, and preserving this variation \textit{is} the goal, this intuition is flipped. We find that many spam annotators are indistinguishable from non-spam annotators, and those that are identifiable are in fact those with very low entropy. Examining the performance of tested methods on completely random vs. completely fixed spam reveals that many methods struggle to identify fixed spam. In particular, methods that correlate with agreement struggle here because a fixed mode response results in relatively high agreement in datasets with substantial variation. These models also struggle on real-world spam in our tested datasets, suggesting that, where preserving variation is paramount, models assuming that spam annotators are more random are not as well suited.

\paragraph{Existing methods work best only when removing few annotators, and distort distributions afterwards.} Tested methods (particularly MACE) are effective at identifying spam annotators for low $n$ (<2-4\% of tested annotators). When more annotators are removed, we see issues across a range of metrics: increased mean absolute error; lower accuracy at spam detection, lower standard deviation, and lower entropy. These issues mean that over-filtering data can lead to labels that do not fully represent the variation in the original distribution.

\paragraph{Detecting spammers vs. detecting low-quality raters.} Since different types of low-quality annotators behave differently, annotators that need to be excluded can exhibit varied behaviors beyond simple patterns such as always selecting the same answer. Consequently, while annotator reliability scoring can often single out spammers showing these stereotypical behaviors, many genuine annotators will be scored similarly to low-quality raters. This result highlights that in addition to the labeling behavior, additional signals should be included in spammer removal. These can be \emph{metadata}, such as when and how much time is spent on annotation \cite{rothwell_controlling_2015} or previous acceptance rates of annotators \cite{difallah_etal_2012_mechanical_cheat}. Similarly, \emph{verifiable test questions} could be used, that is, unambiguous cases where comparison to known answers is possible \cite[gold standard or attention checks,][]{difallah_etal_2012_mechanical_cheat,rothwell_controlling_2015}.

\paragraph{Future work.} Existing methods struggle to distinguish spam from non-spam annotators in contexts where variation in opinion is expected and desirable. This gap highlights the need for spam filtering methods that are robust to variation in labeling behavior.

In addition, the scarcity of available metadata on removed spam data makes it difficult to characterize spammer behavior across a range of contexts. \citet{difallah_etal_2012_mechanical_cheat} highlight a ``need for new benchmarks on which to evaluate and compare existing and novel spam detection techniques for crowdsourcing platforms'' that still persists. Datasets often do not report spam filtering techniques or preserve the spam responses; however, this data is extremely helpful for more fine-grained characterization of spam behavior, especially in complex contexts where variation is expected. Thus, making this data available would be a valuable resource for future research.

\section*{Limitations}

Due to data scarcity, we only used a narrow range of datasets. While the used datasets represent two important use cases where capturing variation matters (AI safety annotations, survey questions), more datasets are needed, especially with different levels of subjectivity, languages and use cases. As such, our results represent only a fraction of relevant scenarios.

Categorizing raters as ``spammers'' is based on varying definitions and procedures: ``gold spammers'' are not ground truth the same way that other data might be. Importantly, self-reported spammer information, where spammers disclose themselves, is largely not even gathered \cite[for an exception, see][]{paun-etal-2018-comparing} and not publicly available. Consequently, the ``gold spammer'' labels used in our study are based on external categorizations. While these are reported to be based on manual checks and multiple data types (labeling behavior, metadata, and attention checks), there remains a risk of wrong categorizations.

We scoped to spam filtering methods that only look at the labeling behavior, given our research question on how this (widely adopted) type of filtering changes the captured variation in labeling. However, there are approaches based on metadata that we could expect to be more effective, perhaps in combination with the evaluated methods using intrinsic metrics based on labeling behavior.

\section*{Acknowledgments}
Matthias Orlikowski and Philipp Cimiano were funded by Volkswagen Foundation as part of the ``Bots Building Bridges (3B)'' project in the ``Artificial Intelligence and the Society of the Future'' programme. Eve Fleisig is partly supported by an NSF GRFP grant.

\bibliography{anthology_0_mod,custom}

\appendix

\section{Computing the CrowdTruth Worker Quality Score}
\label{sec:appendix_crowdtruth}
As highlighted in Section \ref{sec:methods_crowdtruth},  WWA measures how similar a given worker's annotations are to other workers, weighted by the workers' quality and the instances' (or units') ambiguity. WUA measures how much a worker agrees with the aggregate label over all their annotated instances, weighted by the instances' ambiguity. WWA and WUA are roughly computed as follows \cite[ignoring the normalization terms for clarity, full details in][]{CrowdTruth2}:

\[WWA(i) = \sum_{j,u} sim(i,j,u) \cdot WQS(j) \cdot UQS(u)\]

\[WUA(i) = \sum_{u\in units(i)} sim(i,u) \cdot UQS(u)\]

Here, $sim(i,j,u)$ is the cosine similarity between the annotation vectors of workers $i$ and $j$ on an instance $u$. Similarly, $sim(i,u)$ is the cosine similarity between the annotation vector by worker $i$ and the instance vector for instance $u$ (i.e., summed annotation vectors of all other annotators). It is computed over all instances annotated by annotator $i$, denoted $units(i)$. Additionally, $UQS(u)$ measures how much workers agree on an instance $u$ (how ambiguous it is) and is also connected to the workers' quality scores. Due to their inter-dependent nature, the CrowdTruth metrics are re-calculated iteratively until convergence.

\section{Details of Synthetic Spam Results}
\label{sec:appendix}

\begin{figure}
  \centering
  \includegraphics[width=0.85\linewidth]{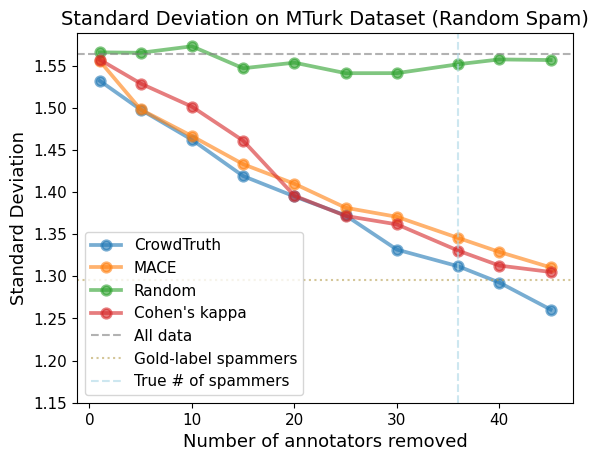}
\includegraphics[width=0.85\linewidth]{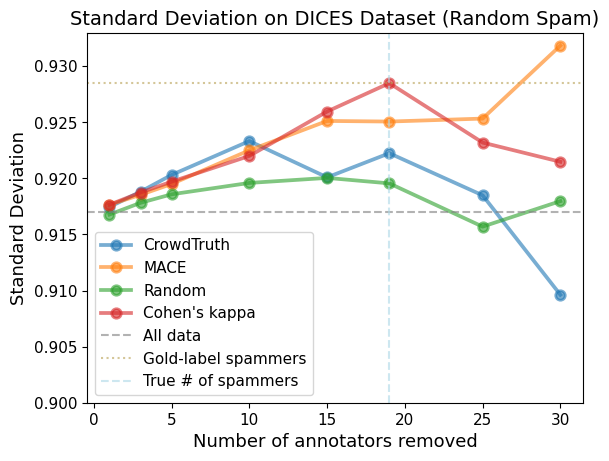}
  \caption{Standard deviation on random spammers.}
  \label{fig:random-stddev}
\end{figure}

\begin{figure}
  \centering
\includegraphics[width=0.85\linewidth]{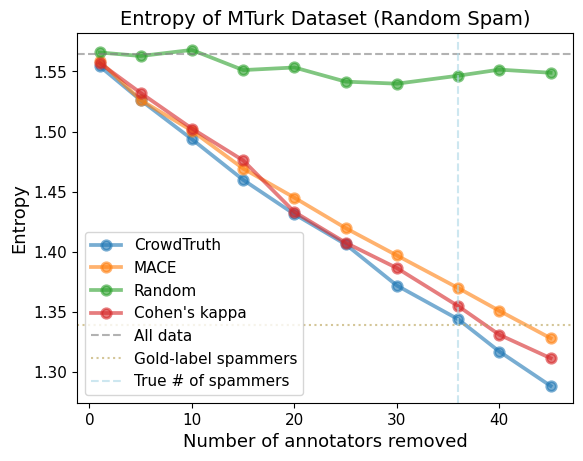}
\includegraphics[width=0.85\linewidth]{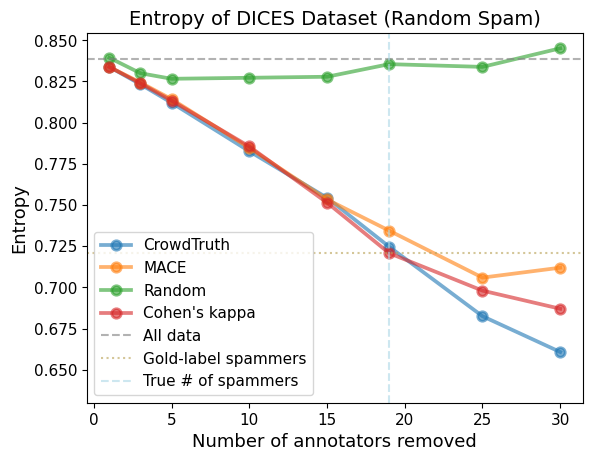}
  \caption{Entropy on random spammers. Entropy generally decreases as more spammers are removed, as expected for accurate spam classification.}
  \label{fig:random-entropy}
\end{figure}

\begin{figure}
  \centering
  \includegraphics[width=0.85\linewidth]{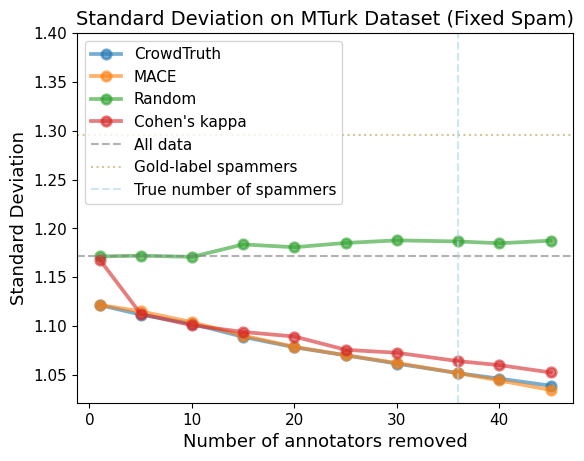}
\includegraphics[width=0.85\linewidth]{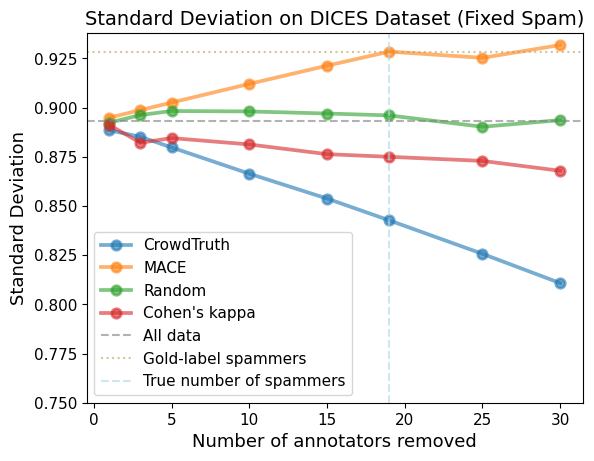}
  \caption{Standard deviation on fixed spammers.}
  \label{fig:fixed-stddev}
\end{figure}

\begin{figure}
  \centering
\includegraphics[width=0.8\linewidth]{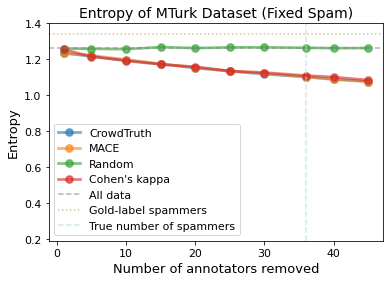}
\includegraphics[width=0.8\linewidth]{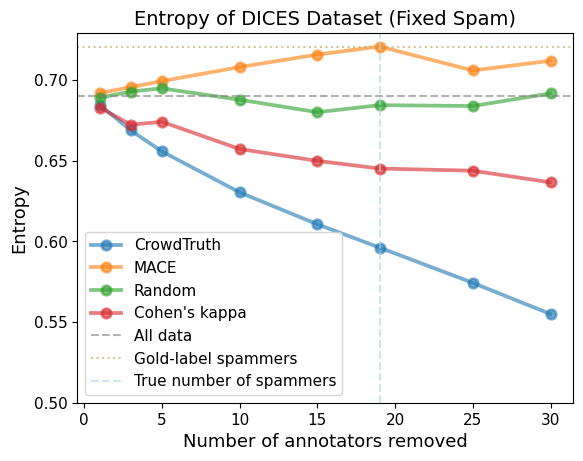}
  \caption{Entropy on fixed spammers.}
  \label{fig:fixed-entropy}
\end{figure}

Standard deviation and entropy for the random and fixed spammers are shown in Figure \ref{fig:random-stddev}, Figure \ref{fig:random-entropy}, Figure \ref{fig:fixed-stddev}, and Figure \ref{fig:fixed-entropy}.

\section{Why does MACE fail to recognize fixed spammers on the MTurk dataset?}
\label{sec:appendix-mace-fixed-spam}
\begin{figure}
  \centering
\includegraphics[width=\linewidth]{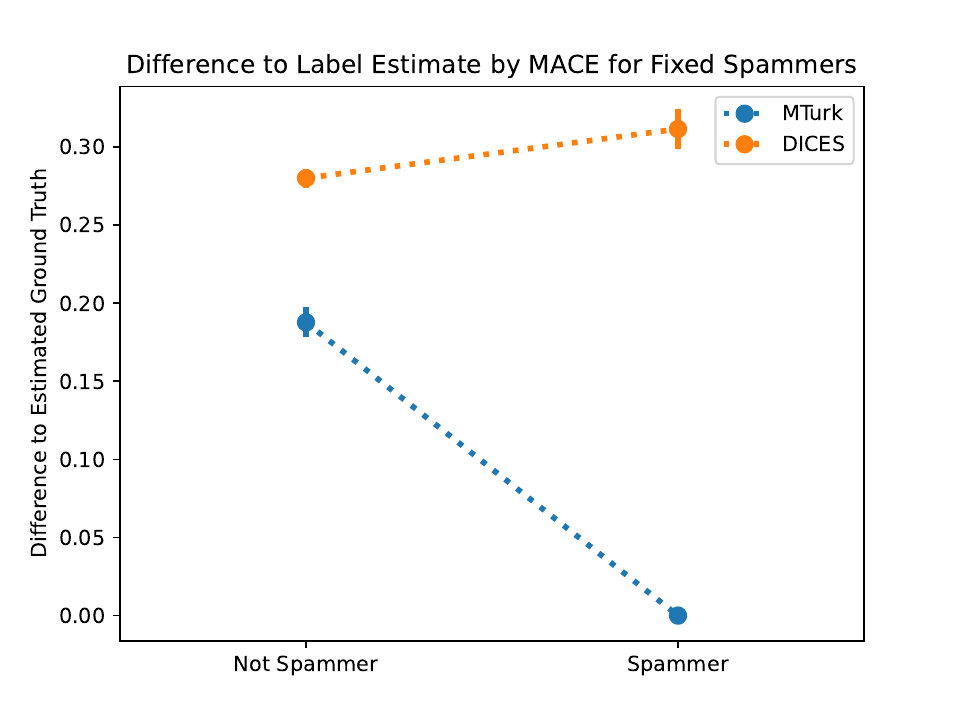}
  \caption{Distance to the ground truth estimated by MACE on fixed spammers vs non-spammers (lower is better). Shows the averaged absolute difference between annotations and the estimated ground truth label. Before averaging, distances are normalized using min-max normalization for each dataset, scaling distances into the range of zero to one.}
  \label{fig:fixed-mace-distance-to-estimate}
\end{figure}

On fixed spammers, who always respond with the mode (the most frequent label in each dataset), MACE gets perfect accuracy on DICES, while on MTurk it performs as poorly as all other methods, failing to reach baseline performance (see Section \ref{sec:results-synthetic-spam}). This result is likely due to the peculiarities of the survey data in the MTurk dataset, where answers follow a normal distribution and the mode is the same for most questions. Here, fixed spammers' seem competent because they always respond with the ground truth as estimated by MACE. Because of this perfect answering behavior of spammers, their average difference to the estimated ground truth is zero, as shown in Figure \ref{fig:fixed-mace-distance-to-estimate}, so that naturally non-spammers are further away from the estimated ground truth, looking less competent to MACE. In contrast, on DICES, which has more varied examples of labeling behavior, non-spammers are on average closer to the estimated ground truth than spammers (see Figure \ref{fig:fixed-mace-distance-to-estimate}).

\section{Dataset Selection Table}
\label{sec:appendix-dataset-selection}

A total of 22 datasets were considered to be included in our study, mostly informed by related work. Table \ref{tab:dataset-selection} lists all of these datasets, including the corresponding references. The table details for each dataset if gold spammer data was collected in principle and if that data was still available. As described in Section \ref{datasets}, we were only able to include two out of these 22 datasets in our experiments.

\onecolumn
    \begin{longtable}{|p{1.2in}|p{1.2in}|p{0.6in}|p{0.6in}|p{2.0in}|}
    \hline
         \textbf{Dataset} &  \textbf{Reference} & \textbf{Gold spammers?} &  \textbf{Included?} &  \textbf{If excluded, why?} \\ \hline
        DICES & \citealt{aroyo2023dices} & Yes & Yes & ~ \\ \hline
        MTurk Survey & \citealt{huang-etal-2023-incorporating} & Yes & Yes & ~ \\ \hline
        MHS corpus & \citealt{sachdeva_measuring_2022} & Yes & No & Raters excluded (details in their paper), but data not available. \\ \hline
        AdultContent3 (''Get Another Label`` datasets) & \citealt{ipeirotis_quality_2010} & No & No & No gold spammers. Experiments in paper use synthetic data and simply report impact on the collected dataset \\ \hline
        HITspam & Discussed in \citealt{ertekin_approximating_2014} & No & No & Despite the name, does not contain spammers. Instead, the task is to judge whether a task on MTurk itself should be considered spam (e.g., because it asks workers to follow a specific social media account). \\ \hline
        EDOS-DOM & \citealt{jiang-etal-2024-examining} & Yes & No & Only one annotator removed after labels were collected. That annotator had annotated only 8 examples (first author vial email). \\ \hline
        Argument Quality & \citealt{mirzakhmedova_are_2024} & No & No & Excludes a number of disagreeing annotations per example. Does not exclude on the level of the annotator. \\ \hline
        MultiPref & \citealt{miranda-etal-2025-hybrid} & No & No & No gold spammers. \\ \hline
        HelpSteer2 & \citealt{wang_helpsteer2_2025} & No & No & No gold spammers. \\ \hline
        CrowdTruth Corpus for Open Domain Relation Extraction & \citealt{dumitrache_false_2017} & No & No & Emailed first author, full data not available anymore. \\ \hline
        AMR / Sentence Similarity Data & \citealt{wein-schneider-2022-accounting} & Yes & No & Only one annotator removed out of three in total. \\ \hline
        Phrase Detectives & \citealt{chamberlain_phrase_2016} & Yes, self-reported & No & Spammer data not available anymore according to authors. \\ \hline
        Crowd-Sourced Preference Tests & \citealt{buchholz_crowdsourcing_2011} & Yes, inferred & No & Data not available anymore according to first author. \\ \hline
        VariErr NLI & \citealt{weber-genzel-etal-2024-varierr} & No & No & Data has annotator IDs and individual decisions plus error judgments (error = no self-validations), but no excluded raters. Not a crowd-sourced study (four annotators). \\ \hline
        Malicious Worker Survey Dataset & \citealt{gadiraju_understanding_2015} & Yes & No & Dataset is not available anymore. \\ \hline
        Dog (Imagenet Subset) & \citealt{5206848} & No & No & No spammer information. Used by \citet{traganitis_identifying_2021}, but only evaluated by accuracy of resulting classifier. \\ \hline
        ImageNetV2 & \citealt{pmlr-v97-recht19a} & No & No & No spammer information. \\ \hline
        Bluebird & \citealt{welinder_multidimensional_2010} & No & No & No spammer information. Used by \citet{traganitis_identifying_2021}, but only evaluated by accuracy of resulting classifier. \\ \hline
        Web & \citealt{zhou_learning_2012} & Unlikely & No & Could not find reference to data. \\ \hline
        WSD & \citealt{snow-etal-2008-cheap} & Unlikely & No & Data not available anymore \\ \hline
        RTE & \citealt{snow-etal-2008-cheap} & Unlikely & No & Data not available anymore  \\ \hline
        TEMP & \citealt{snow-etal-2008-cheap} & Unlikely & No & Data not available anymore  \\ \hline
        POPQUORN & \citealt{pei-jurgens-2023-annotator} & Yes & No & Only a single annotator was removed. \\ \hline
        \caption{Dataset Selection. Shows which datasets where considered and why 20 out of 22 datasets were not included in our study.}
        \label{tab:dataset-selection}
    \end{longtable}

\end{document}